\documentclass[10pt,conference]{IEEEtran}
\IEEEoverridecommandlockouts
\usepackage{caption}
\usepackage{float} 
\usepackage{cite}
\usepackage{url}
\usepackage{booktabs}
\usepackage{amsmath,amssymb,amsfonts}
\usepackage[ruled,linesnumbered]{algorithm2e}
\usepackage{makecell}
\usepackage{diagbox}
\usepackage{graphicx}
\usepackage{subfigure}
\usepackage{textcomp}
\usepackage{xcolor}
\usepackage{color}
\def\BibTeX{{\rm B\kern-.05em{\sc i\kern-.025em b}\kern-.08em
		T\kern-.1667em\lower.7ex\hbox{E}\kern-.125emX}}
	
\usepackage{fancyhdr}
\renewcommand{\headrulewidth}{1pt}

\begin{document}
\title{Multiagent Reinforcement Learning with an Attention Mechanism for Improving Energy Efficiency in LoRa Networks}
\author{\IEEEauthorblockN{
		Xu~Zhang\textsuperscript{$*$},
		Ziqi~Lin\textsuperscript{$*$},
		Shimin~Gong\textsuperscript{$*\dagger$},
		Bo~Gu\textsuperscript{$*\dagger$},
		Dusit~Niyato\textsuperscript{$\ddagger$}
	}
	 \IEEEauthorblockA{
		\textsuperscript{$*$}School of Intelligent Systems Engineering, Sun Yat-sen University, China\\
		\textsuperscript{$\dagger$}Guangdong Provincial Key Laboratory of Fire Science and Intelligent Emergency Technology, China \\
		\textsuperscript{$\ddagger$}School of Computer Science and Engineering, Nanyang Technological University, Singapore \\
	}
	\thanks{This work was supported by the National Key R\&D Program of China under Grant 2020YFB1713800, in part by the National Science Foundation of China (NSFC) under Grand U20A20175, in part by the National Research Foundation, Singapore, and Infocomm Media Development Authority under its Future Communications Research \& Development Programme, DSO National Laboratories under the AI Singapore Programme (AISG Award No: AISG2-RP-2020-019), DesCartes and the Campus for Research Excellence and Technological Enterprise (CREATE) programme, and MOE Tier 1 (RG87/22). \emph{(Corresponding author: Bo Gu.)}}
	\vspace{-1cm}
}

\maketitle
\thispagestyle{fancy}
\fancyhead[C]{This paper has been accepted for publication in IEEE Global Communications Conference (GLOBECOM) 2023.}
\renewcommand{\headrulewidth}{1pt}

\begin{abstract}
Long Range (LoRa) wireless technology, characterized by low power consumption and a long communication range, is regarded as one of the enabling technologies for the Industrial Internet of Things (IIoT). However, as the network scale increases, the energy efficiency (EE) of LoRa networks decreases sharply due to severe packet collisions. To address this issue, it is essential to appropriately assign transmission parameters such as the spreading factor and transmission power for each end device (ED). However, due to the sporadic traffic and low duty cycle of LoRa networks, evaluating the system EE performance under different parameter settings is time-consuming.
Therefore, we first formulate an analytical model to calculate the system EE. On this basis, we propose a transmission parameter allocation algorithm based on multiagent reinforcement learning (MALoRa) with the aim of maximizing the system EE of LoRa networks. Notably, MALoRa employs an attention mechanism to guide each ED to better learn how much ``attention'' should be given to the parameter assignments for relevant EDs when seeking to improve the system EE. Simulation results demonstrate that MALoRa significantly improves the system EE compared with baseline algorithms with an acceptable degradation in packet delivery rate (PDR).
\end{abstract}

\section{Introduction}

Industrial Internet of Things (IIoT), enabling a massive number of end devices (EDs) to communicate and exchange information with each other, is regarded as a key technology for smart grids \cite{smart-grid}, intelligent manufacturing \cite{intelligent-manufacturing}, etc.
These EDs are usually battery-driven and designed to operate for several years without battery replacement, imposing ever higher requirements on their transmission energy efficiency (EE).
Long Range (LoRa) wireless technology, which operates in the licence-free Industrial, Scientific and Medical (ISM) band and provides long-range communication with low power consumption, is very suitable for use in the IIoT \cite{LPWAN}.

The advantages of LoRa technology originate from the design of the physical layer and the media access control (MAC) layer.
In the physical layer, LoRa adopts chirp spread spectrum (CSS) technology to modulate signals to improve the communication range and resistance to interference.
In addition, CSS technology provides LoRa EDs with multiple quasi-orthogonal spreading factors (SFs) for channel multiplexing, thereby increasing the network capacity \cite{scalability-imperfect,Known-and-Unknown}.
In the MAC layer, LoRa Wide Area Network (LoRaWAN) adopts a pure ALOHA mechanism for media access control \cite{LoRaWAN}.
Without carrier sensing, LoRaWAN can achieve low power consumption, especially in small-scale networks.
However, as the network scale increases, the randomness of ALOHA leads to severe packet collisions, thus substantially increasing the energy consumption due to packet retransmission and reducing the system EE \cite{LoRa-beyond-ALOHA}.

Different combinations of the SF and transmission power (TP) for LoRa EDs result in different system performance.
For example, selecting a large SF and TP can increase the transmission range and make the transmission more resilient to noise, but more energy will also be consumed.
Therefore, to adapt to different deployment scenarios and improve the EE performance of LoRa networks, it is essential to optimize the parameter assignments for EDs, which is a challenging task for the following reasons:

\textbf{1) Lack of instantaneous EE performance evaluation:}
Most existing parameter assignment approaches rely on the feedback of  signal-to-noise ratio (SNR) to evaluate the EE performance.
However, LoRa EDs typically operate at a low duty cycle and have a limited data rate, resulting in sporadic packet transmission.
This feature makes it time-consuming to evaluate the EE performance under different parameter settings in real LoRa networks, thus hindering the timely adjustment of the transmission parameters.

\textbf{2) Impact of co-channel interference:}
The co-channel interference in LoRa networks includes co-SF and inter-SF interference.
The former is the interference generated between EDs that select the same SF on the same channel.
The latter is the interference between EDs using different SFs, which is also non-negligible due to the imperfect orthogonality between SFs \cite{scalability-imperfect,Throughput-Analysis}.
The existence of this co-channel interference leads to strong coupling of the transmission parameters, which complicates system EE optimization in LoRa networks.

To address the above-mentioned challenges, we first formulate an analytical model to evaluate the system EE under different parameter settings.
On this basis, we treat each ED as a learning agent and then elaborate a multiagent reinforcement learning algorithm with an integrated attention mechanism (MALoRa) to jointly optimize the SF and TP assignments for LoRa EDs.
Notably, the presented analytical model also facilitates the training of MALoRa by providing immediate EE performance feedback to guide each ED to learn a suitable parameter assignment policy.
The main contributions of this paper can be summarized as follows:

\begin{itemize}
\item \textbf{An analytical model to evaluate the system EE.} We propose an analytical model that considers the effects of channel fading, the capture effect, imperfect orthogonality between SFs and the duty cycle in practical deployment.
Simulation results confirm that the proposed model can accurately evaluate the system EE of LoRa networks under different parameter settings.

\item \textbf{Multiagent-reinforcement-learning-assisted transmission parameter assignment.} To maximize the system EE of LoRa networks, a multiagent reinforcement learning algorithm with an attention mechanism is customized to determine the transmission parameter settings for the EDs. The attention mechanism enables each ED to better learn how much ``attention'' should be given to the parameter assignments for relevant EDs when seeking to improve the system EE.
    
\item \textbf{Significant improvement in system EE performance.} Simulation results demonstrate that MALoRa can learn a good policy to help each ED determine the appropriate transmission parameters, which leads to a significant improvement in system EE compared to existing algorithms.
Furthermore, the effect of the attention mechanism on improving the system EE is confirmed through an ablation study.
\end{itemize}

\section{Problem Formulation}
\begin{figure}[b]
	\vspace{-0.5cm}
	\centering
	\includegraphics[width=76mm]{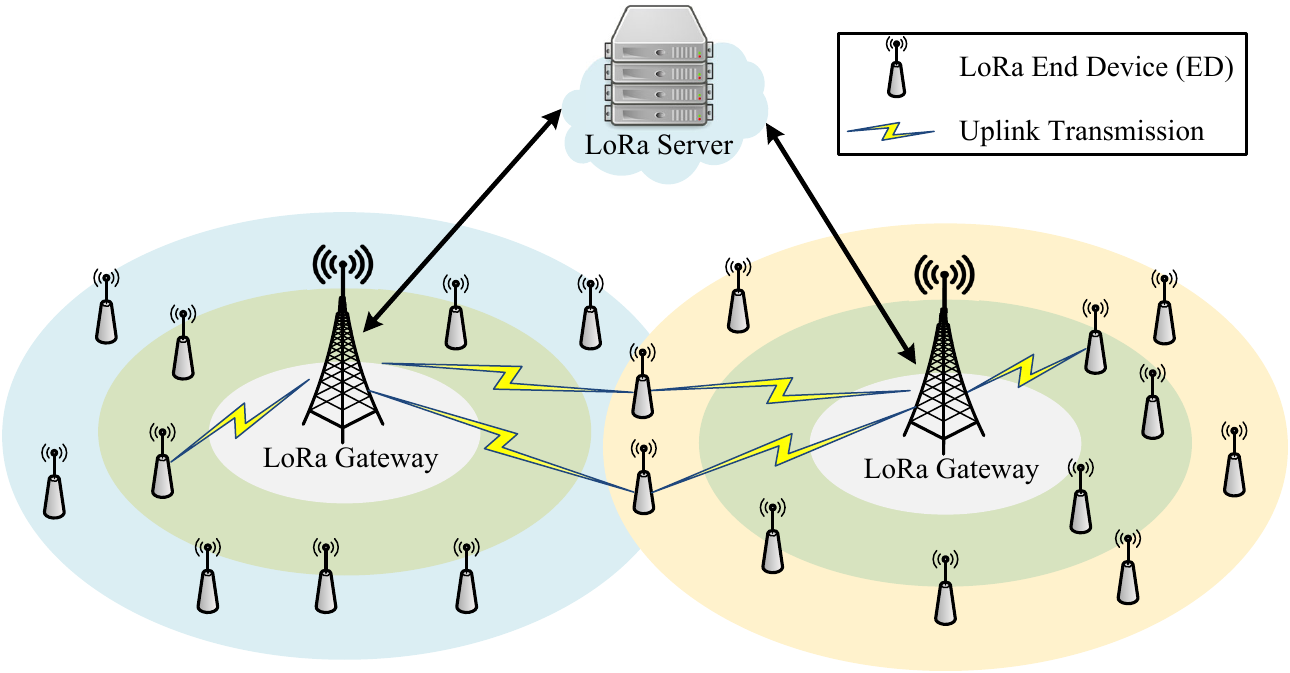}
	\caption{LoRa network resource allocation framework}
	\label{framwork}
	\vspace{-0.25cm}
\end{figure}

Fig. \ref{framwork} shows the LoRa network resource allocation framework.
In this article, we focus on uplink transmission in a LoRa network consisting of $N$ EDs, $K$ gateways and one server.
Specifically, the gateways receive and decode packets from EDs deployed within their transmission range and forward them to the server.
The LoRa EDs communicate with the gateways through $C$ channels and adopt different SFs to share time slots and frequencies within the same channel.
The available TP levels for each ED vary by 2 dB increments between 2 dBm and 16 dBm in the EU863--870 band and thus can be expressed as $\mathcal{P} = \{ 2,4,6,8,10,12,14,16\}$ dBm.
Similarly, the sets of available SFs, channels, EDs and gateways are expressed as $\mathcal{F} = \{ 7,8,9,10,11,12\}$, $\mathcal{C}=\{ 1, \cdots ,C\}$, $\mathcal{N}=\{ 1, \cdots ,N\}$ and $\mathcal{K}=\{ 1, \cdots ,K\}$, respectively.

In accordance with the CSS modulation technique, the LoRa symbols are up-chirp signals that sweep linearly with time over the available bandwidth.
When the SF is set to $f \in \mathcal{F}$, a LoRa symbol can encode $f$ bits of information into a chirp, and the data rate is given by $R_f=f/{T_{\text{sym}}^f}$, where $T_{\text{sym}}^f = {2^f}/{B_c}$ denotes the time duration of a symbol using SF $f$ and ${B_c}$ denotes the bandwidth of channel $c \in \mathcal{C}$.
Selecting a larger SF can increase the communication range and make the transmission more resilient to noise but correspondingly increases the symbol duration and reduces the data rate.

\subsection{Randomness of the LoRa MAC Protocol}

Since LoRa operates in the unlicensed spectrum, the transmission of the EDs should comply with a maximum restriction on the duty cycle.
For example, in Europe, the maximum duty cycle is mandated to be less than 1\% for transmissions in the 868 MHz band.
Under the assumption that the traffic pattern of the LoRa network follows a Poisson distribution \cite{Modeling-Device-level}, the probability of an ED initiating a packet transmission during a given time interval $T$ is defined as $h = 1-{e^{ - \lambda T}}$, where $\lambda$ denotes the packet generation rate.
The experimental results in \cite{Do-LoRa-Scale} show that even when most of the preamble symbols are corrupted by interference, packets can still be decoded as long as the last five preamble symbols are correctly received.
To guarantee successful packet transmission by ED $i$, other interfering EDs should not transmit during the following time interval:
\begin{equation}
	{T'_{ij}} = {T_j} + {T_i} - ({n_{\text{pr}}} - 5)  T_{\text{sym}}^{{f_i}}
\end{equation}
where $T_{\text{sym}}^{{f_i}}$ denotes the symbol duration when using SF $f_i$, ${n_{\text{pr}}}$ denotes the number of preamble symbols, and $T_i$ and $T_j$ represent the time-on-air needed for ED $i$ and the interfering ED $j$, respectively, to transmit a packet.
The time-on-air of ED $i$ can be defined as ${T_i} = T_{\text{pr}}^{{f_i}} + T_{\text{pl}}^{{f_i}}$, where $T_{\text{pr}}^{{f_i}}$ and $T_{\text{pl}}^{{f_i}}$ denote the preamble and payload durations and can be respectively defined as:
\begin{equation}
	T_{\text{pr}}^{{f_i}} = \left( {{n_{\text{pr}}} + 4.25} \right)  T_{\text{sym}}^{{f_i}}, \ \ T_{\text{pl}}^{{f_i}} = n_{\text{pl}}^{{f_i}}  T_{\text{sym}}^{{f_i}}
\end{equation}
Here, ${n_{\text{pl}}^{{f_i}}}$ denotes the number of payload symbols when using SF $f_i$ and can be calculated as:
\begin{equation}
	{n_{\text{pl}}^{{f_i}}} = {8 + \max \left( {\left\lceil {\frac{{8L - 4f_i + 28 + 16}}{{4\left( {f_i - 2DE} \right)}}} \right\rceil   \left( {CR + 4} \right),0} \right)}
\end{equation}
where $L$ represents the payload size of the transmitted packets and $DE$ is the low data rate mode indicator; $DE=1$ if the low data rate mode is enabled, and $DE=0$ otherwise.
Additionally, as the value of $CR$ is varied from 1 to 4, the coding rate $4/(4+CR)$ can be configured as 4/5, 4/6, 4/7 or 4/8.
Therefore, the probability that ED $j$ interferes with ED $i$ becomes:
\begin{equation}\label{h_j}
	{h_{ij}} = 1 - {e^{ - \lambda {{T'_{ij}}}}}
\end{equation}

\subsection{Analytical Model of Transmission Reliability}
According to \cite{EF-LoRa}, the uplink packets transmitted from ED $i$ can be successfully decoded by the LoRa gateways when the following two conditions are satisfied.
First, the received signal strength exceeds the LoRa gateway sensitivity.
Second, the packet is not corrupted due to co-channel interference.
Therefore, the PDR of ED $i$ at gateway $k \in \mathcal{K}$ is defined as:
\begin{equation}\label{pdr_ik}
	PD{R_{ik}} = {\psi _{ik}} {\zeta _{ik}}
\end{equation}
Here, the first factor $\psi _{ik}$ denotes the probability that the received signal strength $RSS_{ik}$ exceeds the gateway sensitivity.
Since the packets transmitted by ED $i$ experience channel fading and path loss before arriving at gateway $k$, the received signal strength is defined as:
\begin{equation}
	RS{S_{ik}} = {p_i} - \overline {PL} ({d_0}) - 10\gamma {\log _{10}}\left( {{d_{ik}}/{d_0}} \right) - {N_\sigma }
\end{equation}
where $p_i$ represents the TP of ED $i$, $\overline {PL} ({d_0})$ is the mean path loss at reference distance $d_0$, $\gamma$ is the path loss exponent, $d_{ik}$ is the distance between ED $i$ and gateway $k$, and $N_\sigma$ is a zero-mean Gaussian random variable with a standard deviation of $\sigma$.
For simplicity, the received signal strength at gateway $k$ can be rewritten as $RSS_{ik} = {z_{ik}} - {N_{ik}}$, where $z_{ik} = p_i - \overline {PL} ({d_0}) - 10\gamma \log_{10} \left( {{d_{ik}}/{d_0}} \right)$ and ${N_{ik}} = \mathcal{N}\left( {0,\sigma  \ne 0} \right)$.
Therefore, $\psi _{ik}$ is calculated as:
\begin{equation}
	\begin{gathered}
		{\psi _{ik}} = \mathbb{P}\left( {RS{S_{ik}} \geqslant {\eta _{{f_i}}}} \right)  = \mathbb{P}\left( {{N_{ik}} \leqslant {z_{ik}} - {\eta _{{f_i}}}} \right) \hfill \\ 
		=  \frac{1}{2} + \frac{1}{2} \text{erf} \left( {\frac{{{z_{ik}} - {\eta _{{f_i}}}}}{{ \sqrt 2 \sigma}}} \right) \hfill \\ 
	\end{gathered} 
\end{equation}
where $\eta _{{f_i}}$ denotes the LoRa gateway sensitivity when the SF is $f_i$ and $\text{erf}\left(  \cdot  \right)$ is the Gauss error function.

The second factor $\zeta _{ik}$ in Eq. (\ref{pdr_ik}) denotes the probability that the packet is not corrupted at gateway $k$ due to co-channel interference.
Due to the randomness of ALOHA, packets from EDs using the same or different SFs may overlap in time when they arrive at gateway $k$.
The capture effect allows packets from an ED suffering such interference to be successfully decoded even if they overlap in time with packets from the interfering EDs, namely, when the signal-to-interference ratio (SIR) of the packet of interest exceeds the threshold.
The SIR thresholds for different pairs of SFs, including co-SF interference $f_i=f_j$ and inter-SF interference $f_i \ne f_j$, are defined in Table \ref{SIR_table}.
The rows of Table \ref{SIR_table} denote the SFs of the ED suffering interference, denoted by ED $i$, and the columns represent the SFs of the interfering ED $j$.
Notably, the co-SF capture threshold is equal to 6 dB for all SFs \cite{Joint-TP-SF, XMAC}.
Given these SIR thresholds, the set of interfering EDs for ED $i$ at gateway $k$ is defined as ${\mathcal{J}_{ik}} = \left\{ {j|RS{S_{ik}} - RS{S_{jk}} < {\omega _{{f_i}{f_j}}}} \right\}$, where $\omega _{{f_i}{f_j}}$ is the SIR threshold for packets using $f_i$ and $f_j$.
For example, when $f_i=8$ and $f_j=10$, if the difference in their received signal strengths at gateway $k$ is below -12 dB, then the packet from ED $i$ is corrupted due to interference.
Therefore, ED $i$ should avoid simultaneous transmission with the EDs in ${\mathcal{J}_{ik}}$ to guarantee successful packet transmission to gateway $k$, and the probability $\zeta _{ik}$ is defined as:
\begin{equation}\label{pro_col}
	\begin{gathered}
		{\zeta _{ik}} = \prod\nolimits_{j \in {\mathcal{J}_{ik}}} {\left( {1 - {h_{ij}}\mathbb{P}\left( {RS{S_{ik}} - RS{S_{jk}} < {\omega _{{f_i}{f_j}}}} \right)} \right)} \hfill \\ 
	\end{gathered} 
\end{equation}
where the probability function $\mathbb{P}\left(  \cdot  \right)$ of Eq. (\ref{pro_col}) can be calculated as:
\begin{equation}
	\begin{gathered}
		\mathbb{P}(RS{S_{ik}} - RS{S_{jk}} < {\omega _{{f_i}{f_j}}}){\text{ }} = \mathbb{P}({N_0} < {\omega _{{f_i}{f_j}}} - ({z_{ik}} - {z_{jk}})) \hfill \\
		= \frac{1}{2} + \frac{1}{2}{\text{erf}}\left( {\frac{{{\omega _{{f_i}{f_j}}} - ({z_{ik}} - {z_{jk}})}}{{2\sqrt 2 \sigma }}} \right) \hfill \\ 
	\end{gathered} 
\end{equation}
where $N_0={N_{jk}} - {N_{ik}}$ denotes the difference between two zero-mean Gaussian distributions with standard deviation $\sigma$, which itself is another zero-mean Gaussian distribution with standard deviation $2\sigma$ \cite{Modeling-Device-level}, i.e., $N_0=\mathcal{N} \left( {0,2\sigma} \right)$.

Given $PDR_{ik}$, the PDR of ED $i$ in a LoRa network with multiple gateways can be defined as:
\begin{equation}
	PD{R_i} = 1 - \prod\nolimits_{k \in \mathcal{K}} {\left( {1 - PD{R_{ik}}} \right)} 
\end{equation}

\begin{table}[t]
	\vspace{0.15cm}
	\centering
	\caption{SIR thresholds in dB for different pairs of SFs \cite{Modeling-Device-level}}
	\setlength{\tabcolsep}{2.5mm}
	\small
	\label{SIR_table}
	\begin{tabular}{|c|c|c|c|c|c|c|}
		\hline
		\diagbox{$f_i$}{$f_j$} & 7   & 8   & 9   & 10  & 11  & 12   \\ \hline
		7                            & 6   & -8  & -9  & -9  & -9  & -9   \\ \hline
		8                            & -11 & 6   & -11 & -12 & -13 & -13  \\ \hline
		9                            & -15 & -13 & 6   & -13 & -14 & -15  \\ \hline
		10                           & -19 & -18 & -17 & 6   & -17 & -18  \\ \hline
		11                           & -22 & -22 & -21 & -20 & 6   & -20  \\ \hline
		12                           & -25 & -25 & -25 & -24 & -23 & 6   \\ \hline
	\end{tabular}
	\vspace{-0.5cm}
\end{table}

\begin{figure*}[!htbp]
	\centering
	\includegraphics[width=150mm]{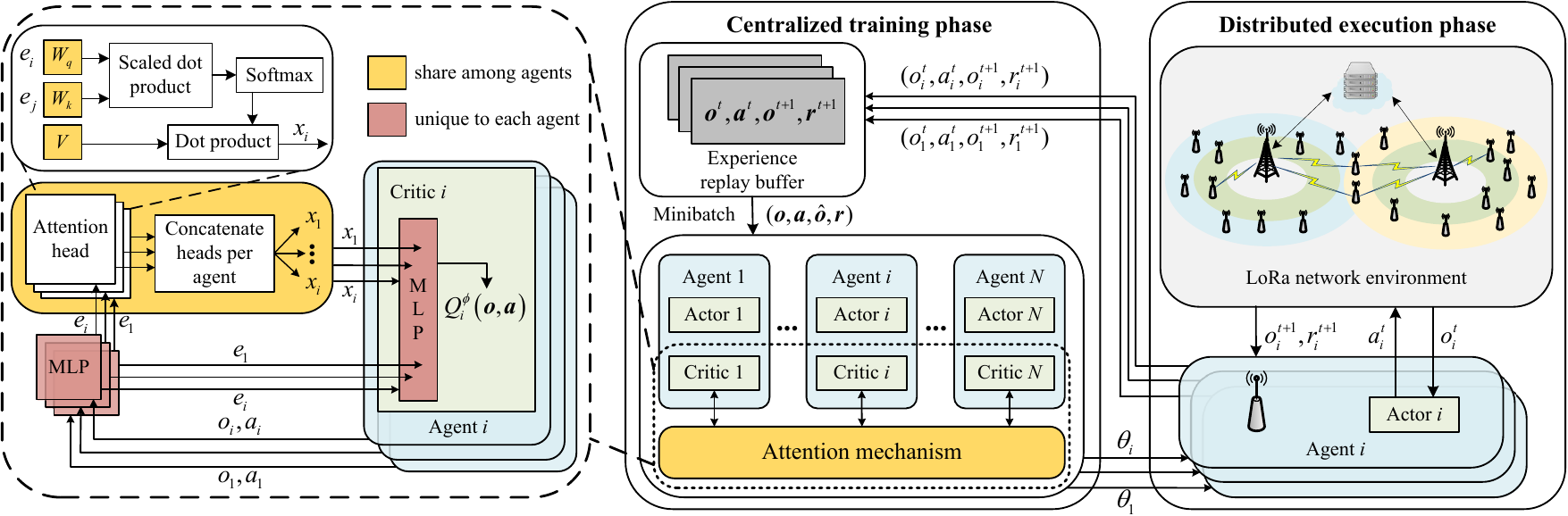}
	\caption{Architecture of the proposed MALoRa algorithm
	}\label{Algorithm}
	\vspace{-0.25cm}
\end{figure*}

\subsection{Optimal Problem Formulation}
The EE of an ED is defined as the total number of bits of information transmitted per unit of energy consumption and is measured in units of bits/mJ.
According to this definition, the EE of ED $i$ can be calculated as:
\begin{equation}
	E{E_i} = \frac{L}{{E_i^{{\text{fix}}}/PD{R_i}}} = \frac{L}{{{e_{{p_i}}}{T_i}/PD{R_i}}}
\end{equation}
where $E_i^{\text{fix}}$ represents the fixed energy consumption for transmitting a packet, which mainly depends on the TP of the ED and the time-on-air related to the SF, and $e_{p_i}$ denotes the energy per unit time required for transmission with TP $p_i$.
Because of the adoption of the ALOHA protocol in LoRaWAN and the presence of interference in LoRa networks, a packet may need to be retransmitted several times before being successfully received by a gateway; the number of such retransmission can be estimated as $1 / PDR_i$ \cite{EF-LoRa}.

Due to the low duty cycle of LoRa networks and the randomness of ALOHA, to simulate a congested channel, we assume that all EDs access the same channel.
Therefore, the main objective under consideration in this article is to jointly optimize the SF and TP assignments for the EDs in a LoRa network so as to maximize the system EE:
\begin{alignat}{12}\label{goal}
	\left( {\bf{P1}} \right) \  \ &\mathop {{\text{max}}}\limits_{\mathcal{F},\mathcal{P}} EE_{\text{sys}} = \sum\nolimits_{i \in \mathcal{N}} {{EE_i}}\\
	\text{s.t.}
	&\ \ X_i^{f_i} \in \left\{ {0,1} \right\},\forall f_i \in \mathcal{F} \tag{\ref{goal}a}\\
	&\ \ {\sum\nolimits_{f_i \in \mathcal{F}} {X_i^{f_i} \leqslant 1} },\forall i \in \mathcal{N} \tag{\ref{goal}b}\\
	&\ \ p_{\text{min}}\leqslant p_i \leqslant {p_{\text{max}}},\forall p_i \in \mathcal{P}  \tag{\ref{goal}c}\\
	&\ \ \ PD{R_i} \geqslant PDR_{\text{th}},\forall i \in \mathcal{N} \tag{\ref{goal}d}
\end{alignat}
Here, Constraints (\ref{goal}a) and (\ref{goal}b) define binary assignment variables $X_i^{f_i}$ to ensure that only one SF can be selected for each ED.
Constraint (\ref{goal}c) states that the available TP levels for each ED range from $p_{\text{min}}$ to $p_{\text{max}}$.
Constraint (\ref{goal}d) specifies the PDR threshold $PDR_{\text{th}}$ to guarantee reliable transmission.

The formulated problem \textbf{P1} is NP-hard due to the coupling of the SF and TP assignments and the nonconvex optimization objective.
To address this problem, the MALoRa algorithm is proposed in this paper to determine the transmission parameter settings for the EDs with the aim of improving the system EE.

\section{MAAC-Based Parameter Allocation Algorithm}

In this section, we design a multi-actor-attention-critic (MAAC) \cite{MAAC} based algorithm called MALoRa to assign transmission parameters to EDs.
To describe how the MAAC method is customized, we first reformulate problem \textbf{P1} as a Markov game and then detail the training process of MALoRa.

\subsection{Markov Decision Process Framework}
Problem \textbf{P1} is first reformulated as a Markov game for the multiagent scenario.
Specifically, each ED acts as a learning agent that aims to learn an appropriate policy for setting its transmission parameters.
The state, action and reward function for each agent are modelled as follows:

\textit{1) State:}
The key features of the LoRa network environment, including the EE and PDR of each agent, should be well captured to ensure that each agent learns an appropriate policy.
The analytical model presented in Section III is used to facilitate the training of MALoRa by providing immediate EE performance feedback to guide each agent to learn a suitable parameter setting policy.
In time slot $t$, the observation of agent $i$ is defined as $o_i^t=\{PDR_i^{t-1},EE_i^{t-1},EE_{\text{sys}}^{t-1}\}$, where $PDR_i^{t-1}$ and $EE_i^{t-1}$ are the PDR and EE of agent $i$ in the previous time slot, respectively, and $EE_{\text{sys}}^{t-1}$ denotes the system EE in the previous time slot.

\textit{2) Action:}
Based on its observation in time slot $t$, agent $i$ chooses appropriate transmission parameters.
The action of agent $i$ is defined as $a_i^t=\{f_i^t, p_i^t\}$, where $f_i^t$ and $p_i^t$ denote the SF and TP allocation decisions of agent $i$, respectively.
Here, the action space of each agent has dimensions of $|\mathcal{F}|\times |\mathcal{P}|$.

\textit{3) Reward Function:}
In this paragraph, we design a reward function that aims to maximize the nonconvex optimization objective given in problem \textbf{P1}.
It is specified that agent $i$ can receive a positive reward when the PDR constraint is satisfied; otherwise, it obtains no reward.
In time slot $t$, the reward of agent $i$ is defined as:
\begin{equation}\label{reward}
	r_i^t = \left\{ {\begin{array}{*{20}{c}}
			{\beta E{E_{{\text{sys}}}} + \left( {1 - \beta } \right){{\overline {EE} }_{ - i}}}&{PD{R_i} \geqslant PD{R_{\text{th}}}} \\ 
			0&{\text{otherwise}} 
	\end{array}} \right.
\end{equation}
where $\beta$ is used to adjust the weight between $EE_{\text{sys}}$ and ${\overline {EE}_{ - i}}$.
${\overline {EE}_{ - i}} = \frac{EE_{\text{sys}}}{N} - \frac{{{EE_{\text{sys}} - EE_i}}}{{N - 1}}$ denotes the effect of agent $i$ on the average EE, where the left fraction is the average EE and the right fraction is the average EE without accounting for agent $i$.
With this reward function, each agent learns the optimal parameter settings by considering not only the EE it achieves but also the impact it causes on the system EE.

\begin{figure*}[!htbp]
	\centering
	\subfigure[]{
		\includegraphics[width=0.3\linewidth]{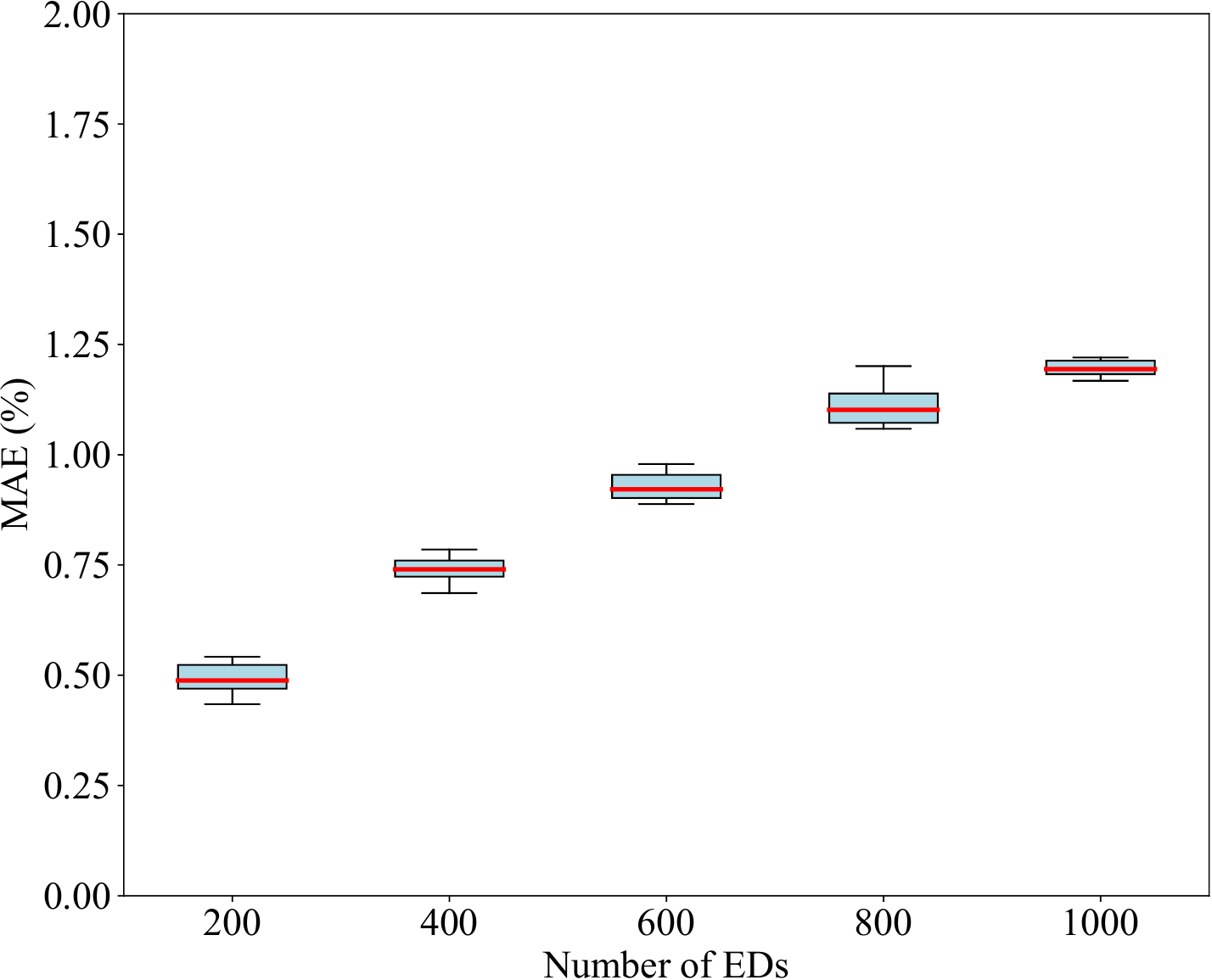}}
	\subfigure[]{
		\includegraphics[width=0.3\linewidth]{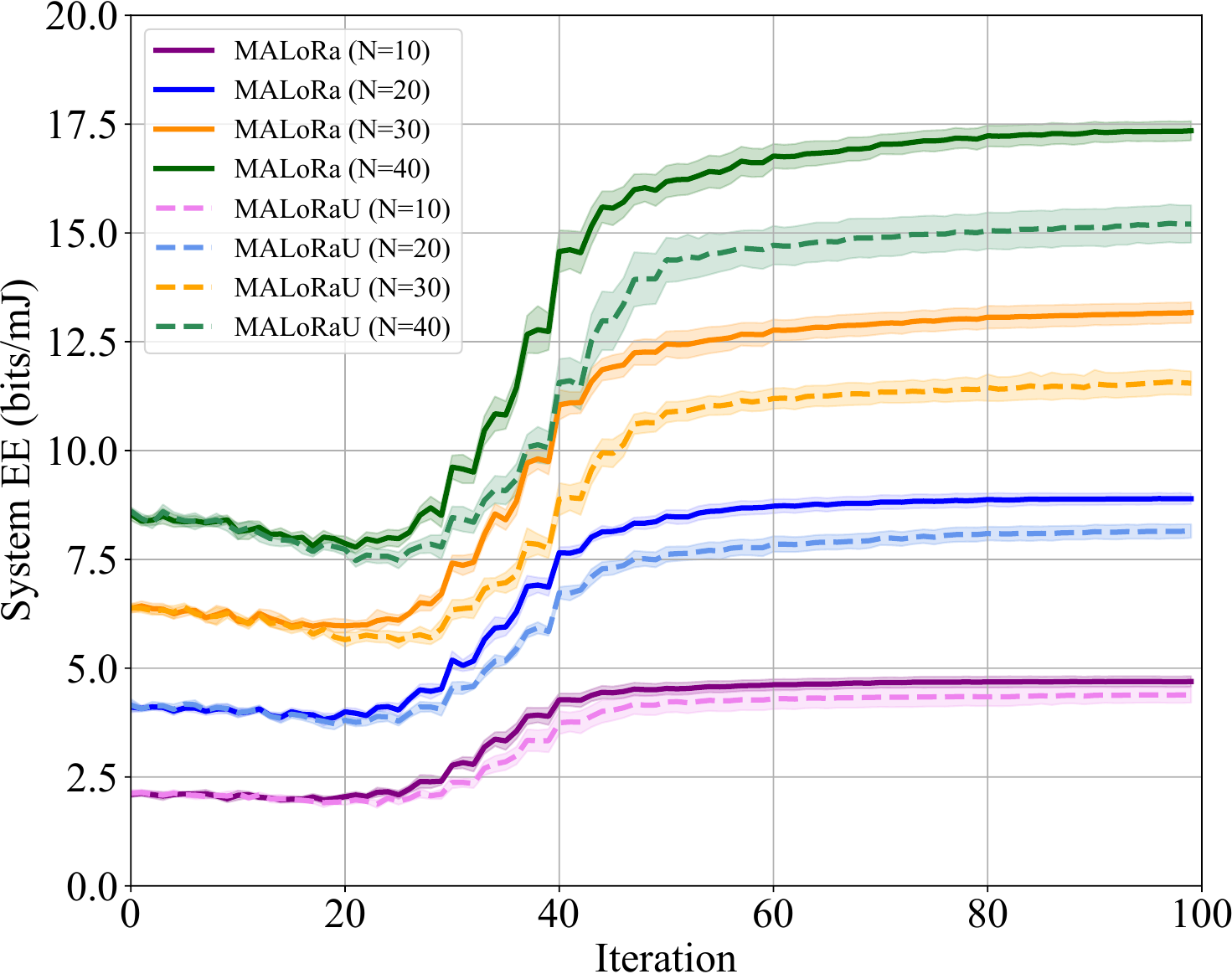}}
	\subfigure[]{
		\includegraphics[width=0.3\linewidth]{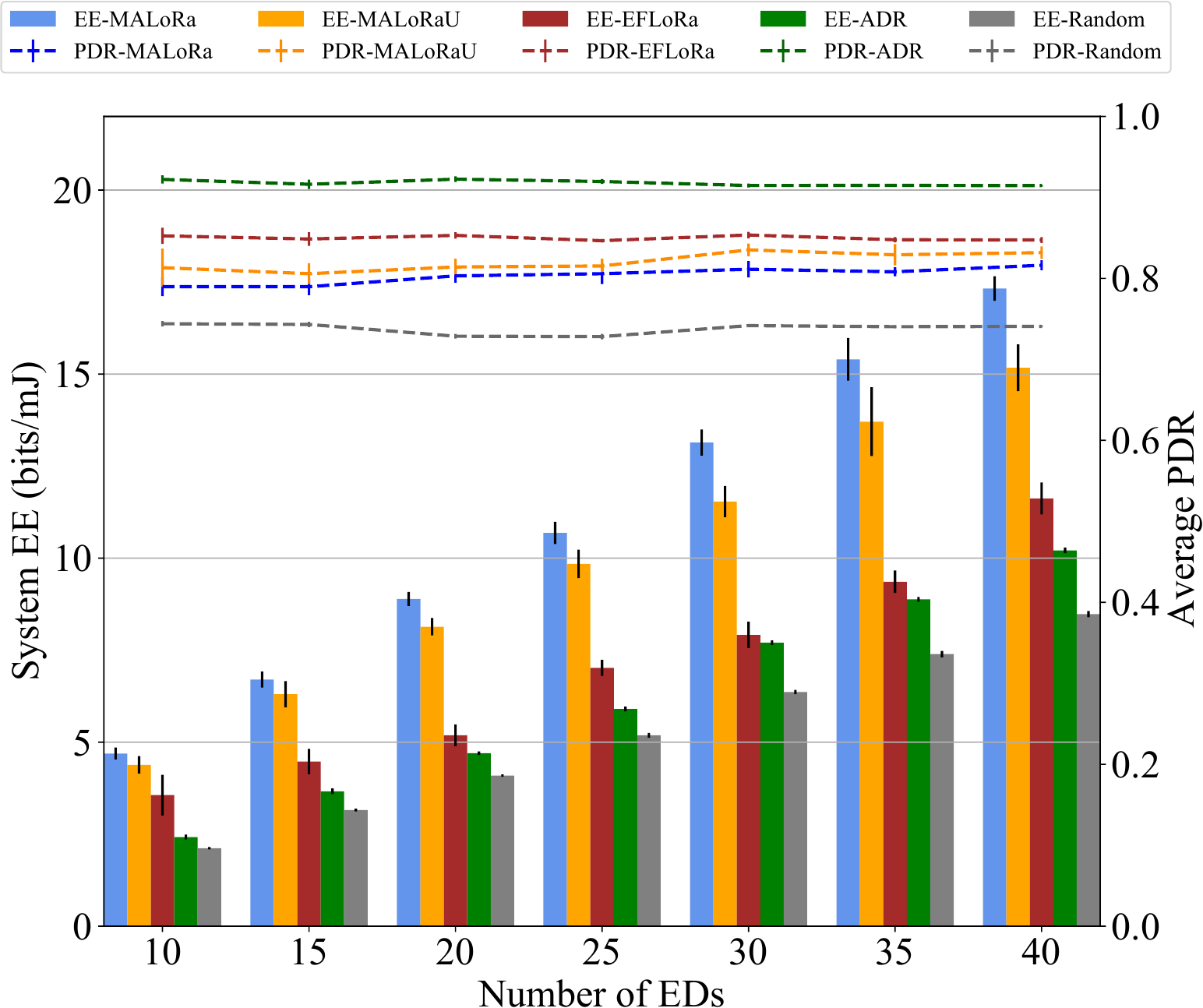}}
	
	\caption{(a) MAE values between the proposed analytical model and the NS-3 simulator for different numbers of EDs (the red solid line represents the median value); (b) System EE versus the number of iterations ($N=10,20,30,40$; $K=4$; $C=1$); (c) Comparison of the system EE and average PDR results achieved with different numbers of EDs under different algorithms ($K=4$; $C=1$; the error bars represent the standard deviation)}
	
	\label{figure_sum}
	\vspace{-0.25cm}
	
\end{figure*}

\subsection{MALoRa Algorithm}
The MAAC algorithm is an RL approach that combines value-based and policy-based approaches based on the actor--critic (AC) algorithm \cite{MAAC}.
Fig. \ref{Algorithm} shows the proposed MALoRa algorithm, which is based on the MAAC algorithm and consists of a centralized training phase and a distributed execution phase.

In the distributed execution phase, each agent $i$ chooses action $a_i^t$ in accordance with its current observation $o_i^t$ by means of its local actor network.
When agent $i$ obtains a new observation $o_i^{t+1}$, a reward $r_i^{t+1}$ is calculated, and then the corresponding transition tuple $(o_i^t, a_i^t, o_i^{t+1}, r_i^{t+1})$ is collected.
Once the transition tuples of all agents have been collected in time slot $t$, the transition tuple $(\boldsymbol{o}^t, \boldsymbol{a}^t, \boldsymbol{o}^{t+1}, \boldsymbol{r}^{t+1})$ is stored in the experience replay buffer $\mathcal{D}$.

In the centralized training phase, a minibatch of tuples $\boldsymbol{B}$ is randomly sampled from the experience replay buffer $\mathcal{D}$ every $T_{\text{update}}$ time slots for training.
Due to the coupling of the transmission parameters, the performance of an agent can be greatly affected by the actions of other agents.
Therefore, each agent needs to attend to the actions and observations of other agents with different weights to obtain the appropriate parameter settings.
To this end, a multihead attention mechanism is adopted in MALoRa.
By selectively attending to transition tuples from $\boldsymbol{B}$ with suitable attention weights, MALoRa can efficiently train a critic network for each agent.
The critic network of agent $i$ receives information from other agents, including their observations and actions, and then incorporates this information to calculate the Q-value function $Q_i^\phi(\boldsymbol{o},\boldsymbol{a})$, which can be represented as:
\begin{equation}
	\begin{aligned}
		Q_i^\phi(\boldsymbol{o},\boldsymbol{a})={F_i}\left( {{g_i}\left( {{o_i},{a_i}} \right),{x_i}} \right)
	\end{aligned}
\end{equation}
where $\phi$ represents the weight parameters of the critic network and $F_i$ and $g_i$ are two-layer and one-layer multilayer perceptron (MLP), respectively.
In addition, $x_i$ denotes the weighted influence of other agents and can be calculated as:
\begin{equation}
	\begin{aligned}
		x_i=\sum\nolimits_{j\in \mathcal{N}, j \ne i} {{\rho _j}h(V {g_j}\left( {{o_j},{a_j}} \right))} 
	\end{aligned}
\end{equation}
where $h$ is the leaky rectified linear unit (ReLU) function and ${g_j}\left( {{o_j},{a_j}} \right)$ is transformed into a ``value'' by a shared matrix $V$.
$\rho_j$ denotes the $j$-th attention weight, which is calculated by comparing the similarity between embeddings $e_j={g_j}\left( {{o_j},{a_j}} \right)$ and $e_i={g_i}\left( {{o_i},{a_i}} \right)$ as follows:
\begin{equation}
	{\rho _j} = \exp (e_j^TW_k^T{W_q}{e_i})/\sum\nolimits_{j = 1}^N {\exp (e_j^TW_k^T{W_q}{e_i})}
\end{equation}
where $e_i$ and $e_j$ are transformed into ``query'' and ``key'' by $W_q$ and $W_k$, respectively.
In the multihead attention mechanism adopted in MALoRa, the set of parameters $(W_k,W_q,V)$ is shared among all critic networks to minimize a joint regression loss function $\mathcal{L}_Q(\phi )$, which is defined as:
\begin{equation}\label{update_critic}
	{\mathcal{L}_Q}\;(\phi ) = \sum\nolimits_{i = 1}^N {{\mathbb{E}_{({\boldsymbol{o}},{\boldsymbol{a}},{\boldsymbol{r}},{\boldsymbol{\hat o}}) \sim \mathcal{D}}}} [{(Q_i^\phi ({\boldsymbol{o}},{\boldsymbol{a}}) - {y_i})^2}]
\end{equation}
where $y_i = {r_i} + \mu {{\mathbb{E}}_{\boldsymbol{\hat a} \sim {\pi _{\bar \theta  } (\boldsymbol{\hat o})}}}\left[ { - \alpha \log \left( {{\pi _{{{\bar \theta }_i}}}\left( {\hat a_i|\hat o_i} \right)} \right) + Q_i^{\bar \phi }\left( {\boldsymbol{\hat o},\boldsymbol{\hat a}} \right)} \right]$ and $\bar{\phi}$ and $\bar{\theta}$ are the weights of the target critic and actor networks, respectively.

To update the individual policies, the gradient ascent algorithm is used. The gradient is defined as:
\begin{equation}\label{update_actor}
	\begin{gathered}
		{\nabla _{{\theta _i}}}J\left( \theta  \right) = {\mathbb{E}_{\boldsymbol{o} \sim \mathcal{D},\boldsymbol{a} \sim \pi }}[{\nabla _{{\theta _i}}}\log \left( {{\pi _{{\theta _i}}}({a_i}|{o_i})} \right) \hfill \\
		( - \alpha \log \left( {{\pi _{{\theta _i}}}({a_i}|{o_i})} \right) + {A_i}(\boldsymbol{o},\boldsymbol{a}))] \hfill \\ 
	\end{gathered}
\end{equation}
where ${A_i}\left( {\boldsymbol{o},\boldsymbol{a}} \right) = Q_i^\phi \left( {\boldsymbol{o},\boldsymbol{a}} \right) - b\left( {\boldsymbol{o},{\boldsymbol{a}_{ - i}}} \right)$ denotes a multiagent advantage function indicating whether the action of agent $i$ will lead to an increase in its expected return, and $b(\boldsymbol{o},{\boldsymbol{a}_{ - i}}) = {\mathbb{E}_{a_i \sim {\pi_i (o_i)}}}\left[ {Q_i^\phi \left( {\boldsymbol{o},\left( {{a_i},{\boldsymbol{a}_{ - i}}} \right)} \right)} \right]$ is the multiagent baseline and ${\boldsymbol{a}_{ - i}} = \boldsymbol{a}\backslash \{ {a_i}\} $ is the actions of all agents except agent $i$.

\section{Performance Analysis}
This section investigates the accuracy of the analytical model and the performance of the MALoRa algorithm through numerical results.
In our simulations, the LoRa EDs are randomly distributed in an area of 8 km$\times$8 km.
The average sending rate $\lambda $ is set to 0.01 $s^{-1}$.
The number of preamble symbols equals to 8 bytes and the coding rate is set to 4/5.
The parameters of the path loss model are determined from measurements taken in the remote neighbourhood \cite{path-loss}, where $\overline {PL} ({d_0})$ = 98.0729 dB, $d_0$ = 40 m, $\gamma$ = 2.1495, and $\sigma$ = 10.0.
The PDR constraint is set to 0.7 and the weight of reward function $\beta$ equals to $1/N$.
The EE of each ED is estimated according to the parameters in \cite{Known-and-Unknown}, which measure the energy consumption of Semtech's SX1276 chip during transmission.

Since the EE of an ED is mainly related to the PDR, we compare the PDRs calculated using the proposed analytical model with those generated by the NS-3 simulator to verify the accuracy of the model.
In our simulations, we assume that each ED randomly initiates an uplink packet transmission.
Each experiment was run in NS-3 for seven days and repeated ten times independently.
The accuracy of the proposed model is measured by the mean absolute error (MAE), which is defined as $\frac{1}{N}\sum\nolimits_{i = 1}^N {\left| {{{\hat x}_i} - {x_i}} \right|}$, where ${\hat x}_i$ and $x_i$ are the PDRs of ED $i$ calculated from the analytical model and obtained using the NS-3, respectively.
The PDR of ED $i$ in NS-3 is defined as the number of packets successfully received by the gateways divided by the total number of packets sent by ED $i$.
Fig. \ref{figure_sum}(a) shows MAE values between the analytical model and NS-3 simulator for each iteration with a 95\% confidence interval.
Each ED in this simulation uses the smallest SF based on the distance to the nearest gateway and the highest TP level of 16 dBm.
The number of EDs varies from 200 to 1000, with all EDs randomly distributed around the four gateways.
Simulation results show that for different numbers of EDs, the proposed analytical model calculates the PDR for each ED with an average error of less than 1.25\%.
As the number of EDs increases, the randomness of the ALOHA and the increase in packet collisions lead to a slight increase in MAE values.

We then evaluate the convergence and the performance of MALoRa under different numbers of EDs.
Due to the low duty cycle of LoRa networks and the randomness of ALOHA, to simulate a congested channel, we assume that all EDs access the same channel.
Fig. \ref{figure_sum}(b) illustrates the system EE versus the number of iterations for varying numbers of EDs.
Each setting was simulated ten times independently, and the results are shown with 95\% confidence intervals.
For comparison, we consider a variant denoted by MALoRaU, which adopts fixed uniform attention weights and thus treats all agents equally.
Through continuously exploring the environment, both MALoRa and MALoRaU can guide the EDs to gradually learn parameter setting policies.
Notably, the system EE gap between MALoRa and MALoRaU is enlarged as the number of EDs increases.
This is not surprising because the uniform attention weights hinder the EDs from obtaining valuable information from other EDs during the training process, and thus, MALoRaU cannot effectively mitigate the co-channel interference caused by the increasing number of EDs.

Fig. \ref{figure_sum}(c) shows the system EE and average PDR results of MALoRa compared with four baseline algorithms (including MALoRaU, EFLoRa \cite{EF-LoRa}, ADR \cite{ADR}, and Random) for different number of EDs.
The simulation results demonstrate that the MALoRa is capable of maximizing the system EE with an acceptable PDR degradation compared to other algorithms.
This is attributed to the attention mechanism employed in MALoRa, which enables each ED to appropriately attend to the parameter assignments of relevant EDs, thus better coping with the coupling of the transmission parameters and reducing co-channel interference to further improve the system EE.
The EFLoRa algorithm achieves limited performance in terms of the system EE because it is designed to solve the max-min problem to achieve EE fairness among the EDs.
The ADR algorithm, on the other hand, tends to select larger SFs and higher TP levels to ensure that the SNR of each ED is above the demodulation floor, which leads to a limited effect in mitigating interference.

\section{Conclusion}
This article aims to improve the system EE in LoRa networks by optimizing the resource allocation for uplink transmission.
We first establish a comprehensive analytical model to calculate the system EE of a LoRa network with multiple gateways.
Simulation results confirm that the proposed model can be used to accurately evaluate the system EE of LoRa networks under different settings compared to the NS-3 simulator.
On this basis, we propose a multiagent reinforcement learning algorithm with an integrated attention mechanism, i.e., MALoRa, to jointly optimize the SF and TP assignments of the EDs.
Simulation results demonstrate that the MALoRa algorithm exhibits good convergence under different settings.
Meanwhile, the attention mechanism employed in MALoRa enables each ED to appropriately attend to the parameter assignments of relevant EDs, thus helping to mitigate interference and further improve the system EE.
In particular, MALoRa significantly improves the system EE compared with four baseline algorithms at the cost of an acceptable degradation in the PDR.

\bibliographystyle{IEEEtran}
\bibliography{refference}

\end{document}